# Developing Neural Network-Based Gaze Control Systems for Social Robots

**Ramtin Tabatabaei[1], Alireza Taheri[1,*]**

[1]Social and Cognitive Robotics Laboratory, Center of Excellence in Design, Robotics, and Automation (CEDRA), Sharif University of Technology, Tehran, Iran

[*]**Corresponding Author:** artaheri@sharif.edu , Tel: +982166165531

**Abstract**

During multi-party interactions, gaze direction is a key indicator of interest and intent, making it essential for social robots to direct their attention appropriately. Understanding the social context is crucial for robots to engage effectively, predict human intentions, and navigate interactions smoothly. This study aims to develop an empirical motion-time pattern for human gaze behavior in various social situations (e.g., entering, leaving, waving, talking, and pointing) using deep neural networks based on participants' data. We created two video clips—one for a computer screen and another for a virtual reality headset—depicting different social scenarios. Data were collected from 30 participants: 15 using an eye-tracker and 15 using an Oculus Quest 1 headset. Deep learning models, specifically Long Short-Term Memory (LSTM) and Transformers, were used to analyze and predict gaze patterns. Our models achieved 60% accuracy in predicting gaze direction in a 2D animation and 65% accuracy in a 3D animation. Then, the best model was implemented onto the Nao robot; and 36 new participants evaluated its performance. The feedback indicated overall satisfaction, with those experienced in robotics rating the models more favorably.

## 1. INTRODUCTION

A variety of kinesics, including speech, bodily movement, pointing, etc., are used in everyday interactions between people to communicate [1]. Among these, gaze is considered as one of the most important factors [2], [3]. Gaze plays a critical role in facilitating a wide range of social interactions and human communication [4]. The direction of gaze indicates a person's interests and preferences, making it an important tool for evaluating mental states, intentions, and predicting human behavior in various social contexts [3], [5], [6], [7]. Additionally, gaze influences a person's sense of presence and can impact how they are perceived [8]. Social robots are specifically designed to perform tasks and engage in scenarios that require close cooperation and interaction with humans [9], [10], [11]. Therefore, besides fulfilling their assigned tasks, social robots must also exhibit respectful social behavior [1], [12], [13], [14]. To be effective in social interactions, robots need to be capable of directing their attention and gaze appropriately in response to social cues, prioritizing the most important person or activity at any given time, especially during multi-party interactions. Hence, a mechanism that can interpret social cues and extract knowledge from visual input is essential for guiding the attention and gaze of social robots [15].

## 2. RELATED WORKS

Research has shown that robot gaze behavior can significantly influence the honesty of human responses during interactions. Pasquali et al. [16] devised a method for lie detection in human-robot interactions by leveraging pupil dilation as an indicator. Their experiment, which involved the iCub humanoid robot in a card game scenario, demonstrated the robot's effectiveness in detecting deception. Similarly, Schellen et al. [17] investigated how robot gaze behavior influences human honesty during interactions. The study found that when a robot makes eye contact after a participant lies, the participant is less likely to lie in subsequent trials.

Advances in gaze control have greatly improved human-robot interaction. Duque-Domingo et al. [18] developed a novel method for gaze control in robotic heads, particularly for scenarios where the robot interacts with multiple people simultaneously. They introduced a competitive neural network that processes various stimuli, such as gaze, speech, and movement, to dynamically determine which person the robot should focus on. Similarly, Lathuilière et al. [9] aimed to enable a robot to autonomously learn and adapt its gaze control strategy, focusing on groups of people using its own audio-visual experiences without relying on external sensors or human supervision. The robot uses a recurrent neural network (RNN) combined with Q-learning, a reinforcement learning technique, to develop an optimal action-selection policy. Lombardi et al. [19] developed a mutual gaze estimation system for the iCub robot, using a learning-based classifier to detect eye contact during human-robot interactions. Additionally, Gillet et al. [20] developed gaze behaviors for robots to balance participation in group conversations using machine learning techniques, comparing imitation learning (IL) and reinforcement learning (RL) approaches, with RL particularly encouraging turn-taking. Finally, Haefflinger et al. [21] used the Furhat robot to test different head and eye movement strategies in group interactions, finding that independent control of the head and eyes, mimicking human behavior, made the robot's interactions feel more natural. Perugia et al. [15] investigated the role of gaze behavior as an implicit indicator of uncanniness and task performance in human-robot interactions, finding that gaze aversion during social chats correlated with perceptions of the robot's uncanniness.

Studying how robots interact with humans while walking has revealed insights into how people focus their attention and perceive social presence. Ise et al. [22] aimed to determine how walking hand in hand with a robot affects the allocation of visual attention and mental workload in humans, compared to walking side by side with the robot without holding hands. Their findings showed that participants looked around their environment more when walking hand in hand with the robot, as opposed to constantly checking the robot's position in the side-by-side condition. Similarly, He et al. [23] investigated how different robot gaze behaviors influence people's perceptions of a robot's social presence in common hallway navigation scenarios, revealing that the preferred gaze behavior is scenario-dependent.

Studies have highlighted the impact of gaze and social cues on enhancing human-robot interaction. Kshirsagar et al. [24] focused on robot gaze behavior during human-robot handovers, finding that a gaze transition from the giver's face to their hand was perceived as likable, anthropomorphic, and effectively conveyed timing and communication cues. Similarly, Terzioğlu et al. [25] demonstrated that adding social cues like gaze behaviors and breathing motions to collaborative robots (cobots) enhances human-robot interaction, further emphasizing the importance of such cues in improving the overall user experience.

Gaze behavior in virtual and augmented reality settings has been explored for its effects on human perceptions and collaboration efficiency. Cuello Mejía et al. [26] examined how different gaze behaviors in virtual agents affect human perceptions during pre-touch interactions in a virtual reality setting, finding that participants preferred complex, two-step gaze behaviors that began with face-looking, as these were perceived as more human-like and natural. Similarly, Holman et al. [27] investigated the role of head and gaze orientation in predicting movement. Both studies found that gaze precedes head orientation and is more accurate in predicting a person's direction, with Holman et al. utilizing virtual reality to track participants' gaze and motion during walking tasks. In a related domain, Weber et al. [28] introduced a method combining augmented reality and eye-tracking to improve human-robot collaboration, enabling real-time robot calibration and object detection using human gaze data. Additionally, Park et al. [29] developed a hands-free human-robot interaction system using eye gazing and head movements in a wearable mixed reality environment, leveraging deep learning for object detection to enhance efficiency in tasks like object manipulation, especially in constrained situations.

To the best of our knowledge, there is a limited number of studies utilizing deep neural networks to model human gaze behavior in social situations, particularly in the context of multi-party conversations. This preliminary exploratory study aims to address this gap and presents our primary contribution. This paper shows the proof of concept for extracting an empirical motion-time pattern for gaze behaviors of human individuals in diverse social situations (based on the gaze behaviors of some participants). The objectives of the paper are as follows: 1) Train deep models to reasonably predict the person to look at based on the ongoing sequence of social situations, 2) Compare the effectiveness of the models derived from both 2D animation and virtual reality headset, and 3) Investigate the performance of the trained models when implemented on a Nao social robot during multi-person social scenarios (i.e. observing some participants' viewpoints obtained from a questionnaire based on the Nao performance).

## 3. METHODOLOGY

The methodology part of the paper is structured into three main sections. The first section delves into two experiments in more detail, focusing on the different video clips for data collection and their corresponding features. It also discusses the social situations portrayed in these video clips and provides insights into the selection of participants for capturing their gaze, along with the setup employed for this purpose. The second section focuses on the architectural design of the neural networks, examining their input and output specifications, as well as the various layers involved. Finally, the last section concentrates on the implementation of these models onto a Nao robot, detailing the necessary setup requirements. Additionally, it outlines the evaluation process for assessing the robot's performance and provides a comprehensive overview of the assessment criteria.

### 3.1. Experimental Setup and the Participants

Our objective is to examine different social situations to determine which ones demand a person's attention the most. To gather data from human participants, we have prepared two video clips. One of them is presented in 2D, and participants view it on a computer; while the other is in 3D, and participants watch it through a virtual reality headset. In the following, we discuss these animations in more depth.

### 3.1.1. 2D Animation

The 2D video clip has dimensions of 1920 × 1080 pixels and a duration of 10 minutes and 40 seconds, a total number of 15342 frames, and is recorded at a speed of 24 frames per second. This video clip comprises various scenarios involving two to four individuals and a stationary box. The animated characters in the video are depicted as men with identical faces and clothing to minimize the impact of facial features and attire on the study's outcomes. The box remains stationary throughout the animation. The reason for placing a box in the 2D animation is that when the characters are pointing, they point at a fixed place. The specific scenario involves four individuals positioned at different angles (-60, -30, 30, and 60 degrees) relative to the user's viewpoint. These individuals are standing and engaged in various activities, with their angles remaining fixed except when they are walking, which allows for angle adjustments as they enter or exit the scene. Additionally, there is a box situated directly in front of the user at a 0-degree angle. The characteristics that each person may have are: 1) Presence or absence, 2) Being near or far, 3) Pointing with the right hand, 4) Waving with the left hand, 5) Talking, 6) Entering or Exiting with high or slow speeds. A total of 128 distinct scenarios are implemented, each lasting approximately 5 seconds and smoothly transitioning to the next mode. At intervals defined by the formula $5+10n$ (e.g., 5, 15, 25 seconds), characters either enter or exit the scene. Additionally, at intervals of $10n$ (e.g., 10, 20, 30 seconds), the activities of the characters change. Our 128 distinct scenarios include: 64 senaroes when there are 3 characters inside the scene; 32 scenarios when there is 2 characters; and 32 scenarios when there is 4 characters. Each activity for each character have two options while these two options are distributed equally. **Figure 1-a** displays a segment of this video clip at the $223^{th}$ second. In the footage, person number 1 is situated in the distance far from the view and is not engaging in any of the mentioned activities. Person number 2 is located near the view and is talking, and pointing. Person number 3 is situated in the distance far from the view and is pointing and waving. Person number 4 is near the view and is waving.

### 3.1.2. 3D Animation

In the 3D video clip, we aim to create a realistic portrayal of social situations by incorporating 2 to 3 individuals placed at various angles (-45, 0, and 45 degrees) and positioned at two different distances from the viewer (i.e., near or far). To ensure that the focus remains on the social context rather than individual faces or features, we have designed all characters in our animation to have the same appearance. The video showcases several social characteristics exhibited by these characters throughout different scenarios, including: 1) Standing (silently or speaking), 2) Moving to the right or left, 3) Moving straight, 4) Waving (silently or speaking), 5) Crossing arms on chest (silently or speaking), 6) Engaging in a conversation with another character, 7) Entering or exiting a scene, and 8) Pointing to a person. When there are 2 people in the scene, we have 12 possible permutations for the standing type; and when there are 3 people, there are 8 possible permutations for the standing type. We have considered all these 20 permutations, and for each permutation, 6 different social situations occur. This results in a total of 120 social situations in our video, seamlessly connected to create a continuous experience. The video is structured to ensure that each character performs actions-standing, waving, crossing arms on the chest, engaging in conversation with another individual, and pointing at someone else-equally across all scenarios. Furthermore, these actions are systematically balanced between being executed silently and accompanied by speech. Each social situation has a duration of 5 seconds before smoothly transitioning into another scenario. By combining these social situations, we have obtained a 10-minute video. Subsequently, we added voiceover to the animation, which will be done using the voicemaker website featuring Matthew's voice. The voices are related to the person's activity. When he is standing, he says, "I am so happy to be with you". During a conversation, the first person says: "I had a math exam today, and it was very difficult". The other party replies: "You have studied a lot for that. I hope you get a good score". While pointing, he says: "Hey, look at him, look at him". When waving, he exclaims: "Hey, look at me over here". And when he crosses his arms, he expresses: "I am so bored; no one pays attention to me". **Figure 1-b** displays a segment of a this video clip at the $107^{th}$ second. In the footage, person number 1 is situated in the distance near the view and is pointing at person 3 and also talking. Person number 2 is located far from the view and is pointing at person 3. Person number 3 is situated in the distance near the view and is waving.

We utilized the powerful and open-source Blender software environment to create both animations, leveraging its versatility and remarkable capabilities in free animation. For the 2D animation, we employed the character named Vincent from Blender Studio, while the 3D animation featured the character Eric obtained from the Free3D website. The 2D animation was created and rendered entirely within Blender; while for the 3D animation, we built character activities in Blender before transitioning to the Unity environment for playback in virtual reality headsets. To streamline the creation process, we incorporated predefined activities from the Mixamo website into our 3D animation.

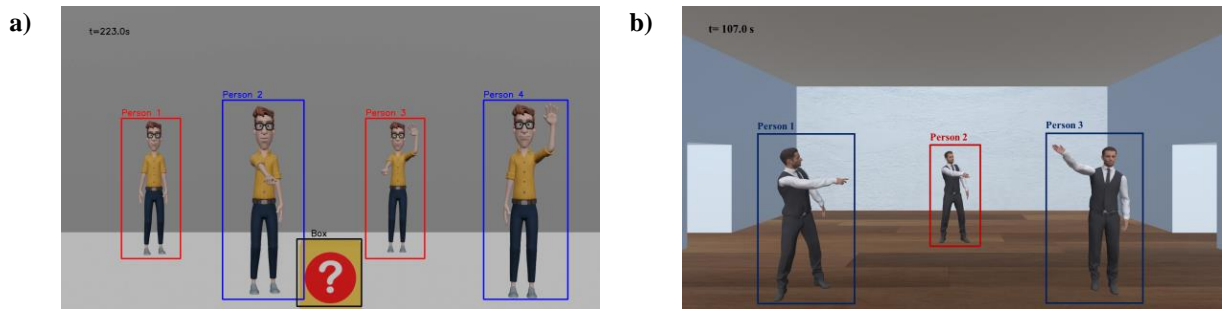

**Figure 1. a)** The position of the people and the box in the 2D video clip in one of the frames, and **b)** The position of the people in the 3D video clip in one of the frames.

### 3.1.3. Data Collection
In the data collection process for the 2D animation, the SR Research EyeLink 1000 Plus device is employed; while for the 3D animation, the Oculus Quest 1 headset is utilized.

### 3.1.3.1 2D Animation
The SR Research EyeLink 1000 Plus device is utilized at the Mowafaghian Comprehensive Rehabilitation Center in Iran to conduct precise eye-tracking tests. This advanced device has a data capture rate of up to 2000 Hz, with a measurement error of approximately 0.15 degrees, ensuring the accuracy and reliability of the collected data. It is equipped with a head holder to keep the participant's head in a fixed position, minimizing unwanted movements and ensuring consistent data capture.

We utilized the SR Research EyeLink 1000 Plus device to record the data at a frequency of 1000 Hz, while the video animation was captured at a frame rate of 24 frames per second. To assign x and y values to each frame, we utilized a sliding window approach, averaging the data points over three consecutive frames: the first and second frames were generated by averaging the first 42 data points, while the third frame was obtained by averaging the last 41 data points within each window. Given factors like blinking, head movement, and device calibration issues, it is inevitable that some data may be missing or deviate from the expected range. Consequently, in order to maintain the accuracy of the experiment's results, we exclude such data and their corresponding frames. The data is organized into 4x7 matrices, where 4 represents the total number of characters, and 7 corresponds to the number of interaction features. These features (which would be used as the input elements for the neural networks) include:

1) Presence of the person (0 or 1)
2) Distance between the person and the user/robot (measured in meters)
3) Waving behavior of the person (0 or 1)
4) Pointing behavior of the person (0 or 1)
5) Talking behavior of the person (0 or 1)
6) Angle of the person relative to the user's/robot's view (in degrees)
7) Movement state of the person (0: standing; 1: entering with low speed; 2: entering with high speed; 3: leaving with low speed; 4: leaving with high speed)

### 3.1.3.2 3D Animation

The Oculus Quest 1 utilizes a sophisticated positional tracking system known as "Inside-Out Tracking" or "Insight Tracking" for 3D animations. This system relies on built-in cameras and sensors to accurately track the user's movements within physical space. While it excels in rotational tracking, it does not track eye movement. Additionally, the Quest 1 incorporates spatial audio technology, providing users with an immersive 3D audio experience that accurately represents sounds originating from various directions and distances within the virtual environment.

For this experiment, we transformed the data recorded by the VR headset into a consistent time step of 0.04 seconds. This adjustment ensured that the data interval was uniform across all data points. The data is organized into 3x6 matrices, where 3 represents the total number of characters, and 6 corresponds to the number of interaction features. These features (which would be used as the input elements for the neural networks) include:

1) Presence of the person (0 or 1)
2) Distance between the person and the user/robot (measured in meters)
3) Characteristic of the person (1: standing; 2: moving left or right; 3: moving forward; 4: waving; 5: crossed arms on chest; 6: conversation; 7: entering and exiting; 8: pointing)
4) Talking behavior of the person (0 or 1)
5) Number of people pointing at him
6) Angle of the person relative to the user's/robot's view (in degrees)

### 3.1.4. Participants

The study involved a total of 30 participants who participated in two experiments. The first experiment focused on the 2D animation and included 15 participants, consisting of 9 males and 6 females. The participants' average age was 25.4 years old, with a standard deviation of 3.14 years. In the second experiment, which focused on the 3D animation, 15 other participants were involved. This group comprised individuals with an average age of 24.0 years old and a standard deviation of 0.65 years. All participants were healthy individuals, free from any diagnosed medical conditions, ensuring that the data collected was representative of the general population. The experimental setups for both experiments are shown in **Figure 2**.

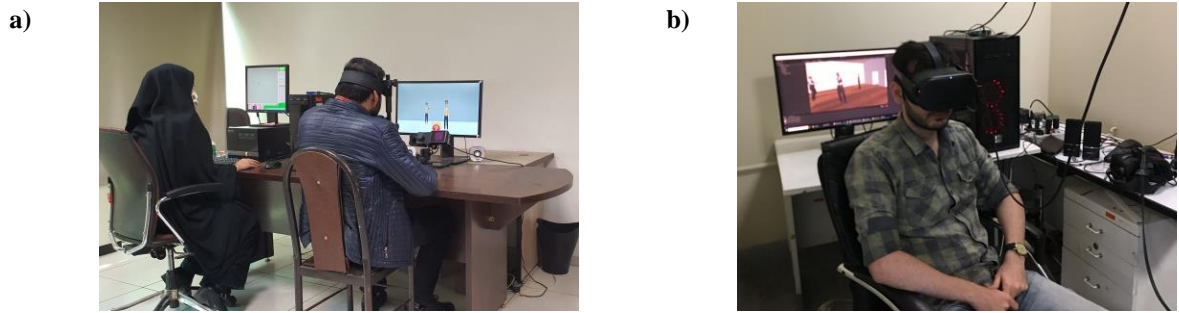

**Figure 2.** The experimental setups of this study: a) eye-tracking system, and b) virtual reality setup.

### 3.2. Deep Neural Networks Architecture

We aim to develop a predictive model to enhance the ability of robots to engage in natural human-robot interactions. Our primary objective is to extract models capable of predicting the optimal gaze directions for a social robot based on the sequence of social situations that have recently occurred. To accomplish this, we have employed two different deep-learning architectures: Long Short-Term Memory (LSTM) and Transformers.

To prepare the input for both networks, we pre-determined the characteristics of each person in every frame. Additionally, we extracted the locations of each person in both experiments. In the first experiment, this was done by defining limitations for the x and y values, while in the second experiment, it was determined based on the amount of rotation angle of the VR glasses during the experiment. In the first experiment, there were five labels for classification: "Person 1", "Person 2", "Person 3", "Person 4", and the "Box." However, in the second experiment, there were only three labels for classification: "Person 1", "Person 2", and "Person 3".

To create a basis for our gaze control system's predictions and detections, we labeled each sequence of frames with the person ID/box of the next frame that the participant was looking at. To ensure the validity of our data, in both experiments, any instance where our participants did not look at any person or the box from the sequence of events was considered as noise data and excluded from our analysis.

Our approach involves providing the model, in both experiments, with one or two additional chances to identify the correct label, as well as varying the frame sequence sizes (i.e., time series before gaze decisions) in one experiment to assess their impact on the model performance. In the first experiment, we considered frame sequence sizes of 0.5 second, 1 second, and 2 seconds (corresponding to 12, 24, and 48 frames, respectively); while in the second experiment, we used a frame sequence size of 1 second (30 frames). A detection attempt of "n" means that the "n" largest model probabilities to identify the correct label are considered. In both experiments, a step size of 1 was used to slide the window on the gaze signals.

The deep learning models in this study were implemented using the Keras library in Python. They were trained using a categorical cross-entropy loss function and optimized with the Adam optimizer, with a learning rate of 0.001 and a batch size of 20. To prevent overfitting, early stopping was applied, terminating the training process if the testing data accuracy did not improve after 10 epochs. The dataset comprised gaze data from 15 participants in each experiment. Model performance was evaluated using 10-fold cross-validation, meaning each model was trained 10 times for each frame sequence size and experiment. During each iteration of cross-validation, nine-tenths of the social situations were used for training, while one-tenth was reserved for testing. The social situations were partitioned such that each set included all social situations as test data exactly once, ensuring that a social situation was only included in the training data (when it was not part of the test data). The average accuracy across all 10 iterations was calculated for each experiment, providing a comprehensive evaluation of model performance. The training of the neural networks was conducted in the Google Colab environment, with each epoch taking approximately one minute. The following sections will delve into the architecture used for each approach in greater detail.

### 3.2.1. Modelling with LSTM algorithm

The LSTM algorithm is highly effective at capturing time-related patterns in sequential data. In the proposed model, we utilized two LSTM layers, each consisting of 64 units activated by the hyperbolic tangent function (tanh). **Figure 3** demonstrates how the input data progresses through these two LSTM layers, and the output of the second LSTM layer is then directed to a softmax layer with either 3 or 5 neurons for classification purposes.

The neural network takes input in the shape of (n, m, L), where n represents the total number of sample data, m represents the size of frame sequence, and L represents the number of features for the maximum number of people in the scene. In the first experiment, we had a training dataset of 160,000 samples and a test dataset of 19,000 samples; while in the second experiment, we had 190,000 training samples and 20,000 test samples. The frame sequence sizes in the first experiment varied between 12, 24, and 48 frames (equivalent to 0.5s, 1s, and 2s), while the frame sequence size in the second experiment was fixed at 30 frames (1s) (It should be noted that based on the observed results from the first experiment, which are presented in the results section of the paper, the effect of the frame sequence size on the accuracy of the models is very minimal. Therefore, we decided to keep it fixed in the second experiment). In the first experiment, there were 4 persons, each with 7 features, resulting in L being equal to 28. In the second experiment, there were 3 persons, each with 6 features, making L equal to 18. The model parameters for the first experiment were 57,157, and for the second experiment, they were 54,467.

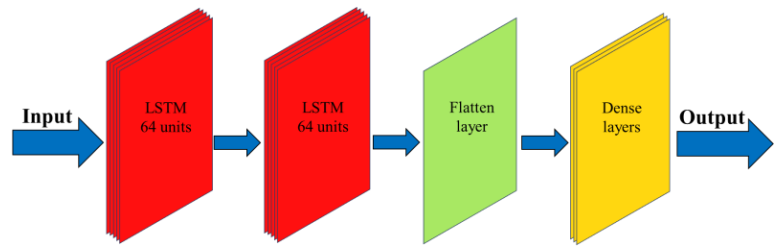

**Figure 3.** Neural network architecture for the used LSTM architecture.

### 3.2.2. Modelling with Transformers algorithm

The transformer architecture has revolutionized natural language processing by introducing self-attention mechanisms for efficient input sequence processing [30]. Unlike traditional recurrent neural networks, transformers can effectively capture long-range dependencies by attending to all input tokens simultaneously. Our system takes the input as normalized numbers in vectors, eliminating the need for Input Embedding. We incorporate learned positional encoding with an input dimension equal to the size of the frame sequence and an output dimension equal to the total number of features. In our system, we employ two Encoder blocks, each comprising a MultiHeadAttention component with 2 attention heads, where the size of each attention head matches the total number of features. Additionally, the Encoder block incorporates a position-wise feed-forward network consisting of two dense layers that utilize the swish activation function. These layers have dimensions of 1024 and the total number of features, respectively. To preserve important details and context from the original sequence, we merge the outputs of the Encoder block with the original inputs by perform an element-wise addition operation. Subsequently, the resulting vector undergoes GlobalMaxPooling1D, which extracts the maximum value from each feature dimension, effectively reducing the sequence length. Finally, the processed vector is passed through a dense layer with dimensions matching the number of labels (e.g., 3 or 5), generating the final predictions. The model architecture is illustrated in **Figure 4**.

The input data (and its dimension/shape) for the transformer network is exactly the same as our LSTM network. In the first experiment, the number of training parameters is 130,433; and for the second experiment, it is 81,989.

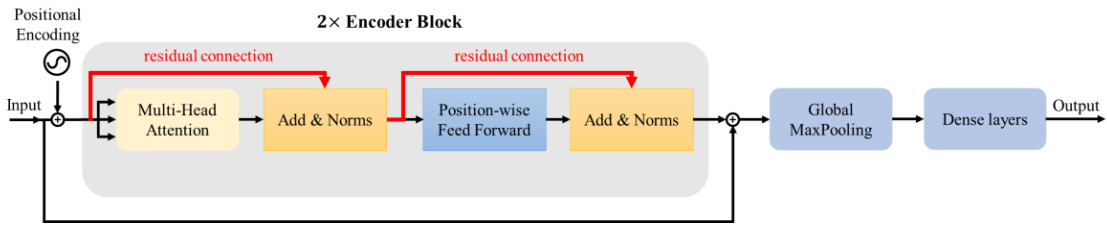

**Figure 4.** Neural network architecture diagram of the Transformers architecture.

### 3.2.3. Implementation and evaluation of the best gaze model on the social robot

In this section, we describe the methodology for deploying our machine learning models on a Nao robot. In order to assess the practical performance of our models, we administer a questionnaire to participants, seeking their feedback on various aspects of the robot's interaction. We utilize the Glasgow Coma Scale to gauge participants' attitudes and sentiments towards the robot, encompassing factors such as satisfaction with the interaction, perception of the robot's human-like qualities, evaluation of its intelligence, and comprehension of human actions.

To accurately determine the position, sound direction, and actions of individuals in front of the robot, we incorporated the Kinect 2 sensor. This technology enables us to achieve precise person detection and landmark localization. By leveraging specific thresholds and utilizing kinematics energy, we are able to detect and track the actions and movements of each individual effectively. Moreover, Kinect 2's high-quality microphone arrays facilitate the detection of sound direction accurately. Through the integration of Kinect 2's advanced features with our neural network designs, our robot exhibits improved understanding and responsiveness to the individuals in its environment. The implemented network operates on a Nao robot, which is widely recognized as a visually appealing humanoid robot. Its sleek white design and rounded features contribute to a friendly and approachable appearance. Developed by Softbank Robotics in France, the Nao robot represents years of research and development in the field of robotics, particularly in the realm of social robotics. One of Nao's notable features is its ability to rotate its head at different speeds, providing greater flexibility in its interactions with humans. This capability allows for more accurate tracking of movements and enables the robot to respond promptly to changes in its environment. Alongside this impressive feature, the Nao robot possesses a range of capabilities that make it highly versatile and effective for diverse applications.

In order to evaluate the performance of the models on the Nao robot, we recorded two comparable 2-minute videos. The experimental setup involved two to three individuals standing in front of the Nao robot, engaging in activities such as entering, standing, moving, waving, pointing, crossing arms, and leaving (similar to the features considered in the animations). These videos were then shared with a group of 36 participants through social media. The participants consisted of 23 males and 13 females, with an average age of 24.34 years old and a standard deviation of 3.83. They were divided into two groups: 12 experts (participants with robotics experience), and 24 novices (participants with no prior knowledge or lacking in-depth knowledge of robots).

After watching the videos, the participants were requested to complete a questionnaire designed to evaluate their emotions and attitudes towards the robot and its performances. The questionnaire utilized a 5-point likert scale consisted of a series of statements designed to evaluate the performance of the best model (highest accuracy) trained on the robot using both the 2D animation dataset and the virtual reality dataset. An anonymous questionnaire was developed, incorporating variables such as age, gender, participants' familiarity with robotics, as well as a set of questions inspired by the UTAUT [31] and a questionnaire from a similar previous study [32]. Participants were asked to indicate their level of agreement with ten statements presented in **Table 1**. A five-point Likert scale was employed, with verbal anchors ranging from 'totally disagree' to 'totally agree'. The randomization of the video and statement order was implemented to ensure impartiality and minimize potential biases (i.e., counterbalanced condition). This counterbalance condition aimed to create an unbiased evaluation environment, enhancing the reliability of the assessment. Further details and results are provided in the Results section. Two-way mixed Anova analysis will be applied on the scores of the questionnaire's statements. We considered 'Models' (with two levels: 2D and 3D animation) and 'Robotics Knowledge' (with two levels: experts and novices in robotics) as the two main independent factors in our ANOVA statistical test. 'Models' was treated as a within-subjects factor, while 'Robotics Knowledge' was treated as a between-subjects factor.

**Table 1.** Evaluation questions to be asked of participants in the experiment to assess the performance of the robot.

| *Statements* |
| --- |

| | |
|---|---|
| Statement 1: | I feel satisfied with the interaction of the robot in the video. |
| Statement 2: | I'm satisfied with the robot's coordination with people's movements in the video. |
| Statement 3: | I believe the robot in the video is a great social companion. |
| Statement 4: | The robot's interaction with the people in the video felt remarkably human-like to me. |
| Statement 5: | The robot in the video appears so lifelike that I can easily imagine it as a living being. |
| Statement 6: | The robot in the video pays good attention to the people around it. |
| Statement 7: | The robot behaved intelligently in the video. |
| Statement 8: | In my opinion, the robot demonstrates a good understanding of the people in the video. |
| Statement 9: | The robot in the video responded adeptly to the actions of the people. |
| Statement 10: | The robot in the video exhibited a solid understanding of the actions of the people. |

## 4. RESULTS

This section presents and discusses the results obtained from the two algorithms employed in the study, as well as the results of the questionnaire. The results are divided into four main parts. Firstly, we compare the accuracy of the models extracted from the 2D animation on both the training and test data using both algorithms. Secondly, we demonstrate the model's accuracies on the dataset collected from the virtual reality headsets. Thirdly, we demonstrate the models' performance on some selected frames from the video clips. Lastly, we present the results of a questionnaire that assesses the performance scores of our models in real-life scenarios, and compare the models' effectiveness with similar works in the field.

### 4.1. Results of the first experiment (2D Animation)

**Figure 5** illustrates the accuracy of two different neural network architectures and the models trained using the 2D animation dataset. The graphs display the accuracy of the models based on three different frame sequence sizes and the number of detection attempts. This approach provides the models with an increased opportunity to correctly identify the labels.

**Figure 5-a** and **Figure 5-b** depict the accuracy of the models on the training data. The first figure focuses on the accuracy of the models using LSTM, while the second figure highlights the accuracy when using Transformers. The results show that both architectures achieved similar results in each detection attempt and frame sequence size (with an accuracy difference of less than 0.80%). Additionally, the results indicate that the frame sequence size does not significantly affect the accuracy of the models. When using the LSTM architecture, the accuracy of the model with a single detection attempt and a frame sequence size of 12 is 66.2% (±0.8%), and with a frame sequence size of 48, the accuracy is 67.0% (±0.9%). As the number of detection attempts increases to 2, the accuracy of the LSTM model with a frame sequence size of 24 improves to 89.1% (±0.5%). Similarly, when the number of detection attempts increases to 3, the accuracy of the LSTM model with a frame sequence size of 24 further increases to 97.6% (±0.2%).

**Figure 5-c** and **Figure 5-d** showcase the accuracy of the models on the test data, their ability to predict the correct person/box to focus on in unseen social situations. The first figure focuses on the accuracy of the models using LSTM, while the second figure highlights the accuracy when using Transformers. The results demonstrate that both architectures achieved similar results in each detection attempt and frame sequence size (with an accuracy difference of less than 1.45%). Furthermore, the findings suggest that the frame sequence size does not significantly impact the accuracy of the models. When utilizing the LSTM architecture, the accuracy of the model with a single detection attempt and a frame sequence size of 12 is 61.5% (±3.9%), and with a frame sequence size of 48, the accuracy is 58.8% (±4.8%). As the number of detection attempts increases to 2, the accuracy of the LSTM model with a frame sequence size of 24 improves to 81.3% (±2.8%). Similarly, when the number of detection attempts increases to 3, the accuracy of the LSTM model with a frame sequence size of 24 further rises to 93.8% (±2.0%).

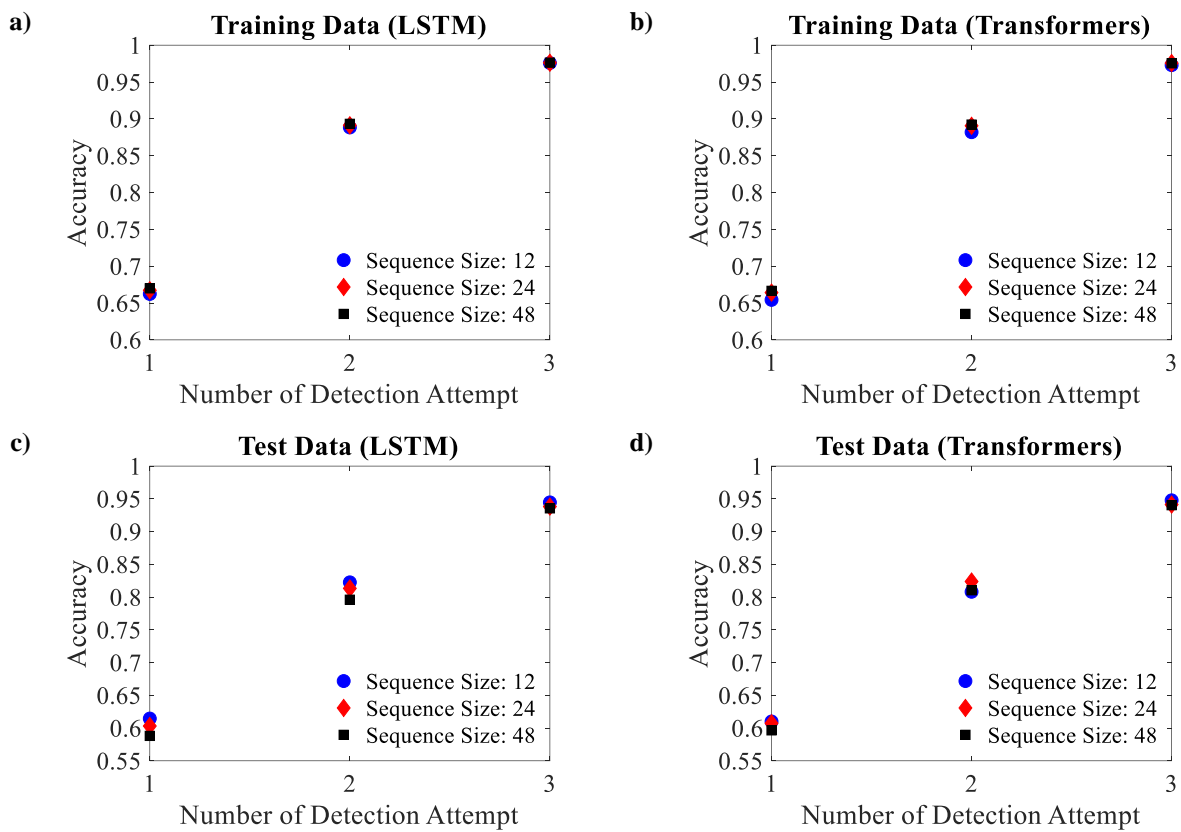

**Figure 5.** The accuracy of the model using LSTM and Transformers algorithms when the number of labels is 5 (i.e. 4 persons and the box) on **a,b)** the training data, and **c,d)** the test data.

### 4.2. Results of the second experiment (Virtual reality)

**Figure 6** shows the accuracy of two different neural network architectures and the models trained using a dataset collected from a virtual reality headset. The accuracy of the models is depicted based on the number of detection attempts. This approach provides the models with an increased opportunity to correctly identify labels. The frame sequence size in this experiment is set to 30 frames. The accuracies reported are the average results from 10 different runs of K-Fold cross-validation.

**Figure 6-a** focuses on the accuracy of the models on the training data, specifically using LSTM and Transformers architectures. The results demonstrate that both architectures achieved similar accuracy levels in each detection attempt, with a difference of less than 0.04%. For the LSTM architecture, the model achieved an accuracy of 74.7% (±0.9%)

with a single detection attempt, while the Transformers model achieved 74.8% (±0.7%). As the number of detection attempts increased to 2, the accuracy of the LSTM model improved to 96.8% (±0.3%).

Moving on to **Figure 6-b**, it showcases the accuracy of the models on the test data, evaluating their ability to predict the correct person to focus on in unseen social situations. Again, LSTM and Transformers architectures are compared. The results indicate that both architectures achieved similar accuracy levels in each detection attempt, with a difference of less than 1.89%. With a single detection attempt, the LSTM model achieved an accuracy of 64.5% (±9.1%), while the Transformers model achieved 63.9% (±9.0%). As the number of detection attempts increased to 2, the accuracy of the LSTM model improved to 91.8% (±8.9%).

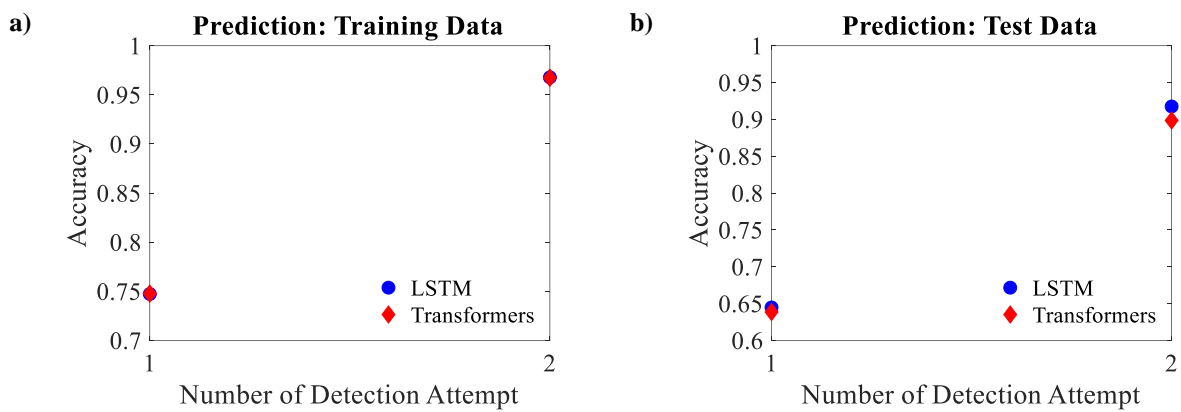

**Figure 6.** The accuracy of the model using LSTM and Transformers algorithms when the number of labels is 3 (i.e. 3 persons) on **a)** the training data and **b)** the test data.

### 4.3. Result of the model on some of the video clip's frames

To provide a visual representation of our model's performance, we have included the results of our LSTM models from both experiments in the video clips (i.e., the best models), showcasing their ability to predict the correct person to focus on in untrained social situations. **Figure 7** displays some frames from the video clip with accurate/correct predictions (based on the test data). In these figures, the small red circles represent the gaze locations of each of the 15 participants in each experiment, while the small white star indicates the location the model predicts should be looked at.

**Figure 7-a** captures the 22$^{nd}$ second of the 2D animation video clip, where person number 1 and person number 4 are standing far apart, person number 2 and person number 3 are nearby and pointing and waving, and person number 3 and person number 4 are engaged in a conversation. It can be observed that most participants looked at person number 3, and the model accurately predicts that the focus should be on person number 3 as well. **Figure 7-b** presents the 113$^{th}$ second of the 2D animation video clip. In this frame, person number 1 is nearby, waving, and talking, person number 2 is far away, talking, waving, and pointing, while person number 3 is nearby and not involved in any of the mentioned activities. As shown, the majority of participants looked at person number 2, and the model correctly predicts that the focus should be on him.

**Figure 7-c** displays the 68$^{th}$ second of the Virtual Reality video clip, where persons number 1 and 2 are engaged in a conversation nearby, and person number 3 enters from a distance. It is evident that most participants looked at persons number 1 and 2, and the model correctly predicts that the focus should be on person number 1 based on this social situation. **Figure 7-d** illustrates the 340$^{th}$ second of the video clip. Person number 1 is far away and waving, while person number 3 is nearby with crossed arms. As observed, most participants looked at person number 1, and the model accurately predicts that the focus should be on him.

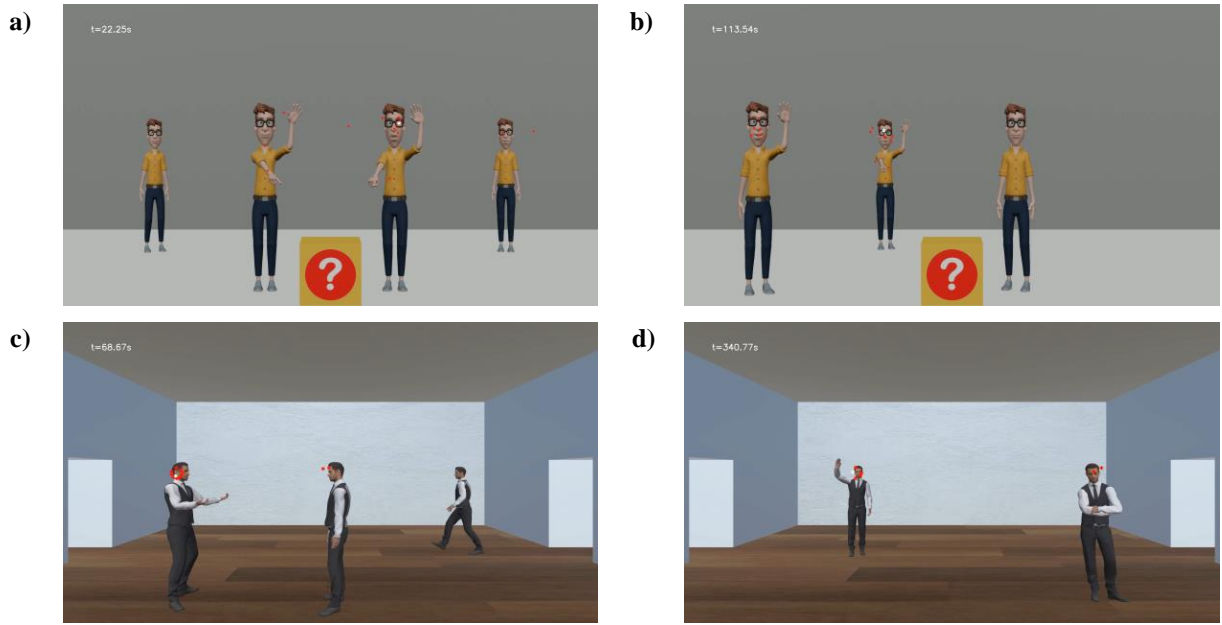

**Figure 7.** Some examples of the participants' look and the model's correct predictions in the 2D or 3D video clips.

**4.4. Implementation and Performance Evaluation of the Best Neural Network Model on a Robotic System**

In the upcoming section, we explain the implementation of our models on the Nao robot by utilizing data from both 2D animation and virtual virtuality datasets. Our primary focus is on using the most accurate LSTM model available. Subsequently, we provide a detailed discussion of the evaluations given by the participants regarding the robot's performance in real-world settings. We examine their feedback and responses regarding their satisfaction with the robot's interaction, the robot's human-like qualities, its intelligence, and its understanding of human actions in subsequent sections, offering further insights. Then, we conducted a two-way mixed ANOVA test with a confidence level of 95% and with the factors of the model and participants' knowledge about robots.

As mentioned earlier, we selected the Nao robot from the available options in the laboratory as the ideal choice for implementing our models. This decision was primarily based on the robot's user-friendly nature and visually appealing design. However, it is important to note that the chosen Nao robot lacks a depth sensor. To overcome this limitation, we incorporated the Kinect 2 device, which serves as a reliable depth sensor and also has the capability to detect landmarks of individuals within its field of view. Additionally, the Kinect 2 is equipped with an array of microphones, allowing accurate sound direction detection. In **Figure 8**, we successfully demonstrate the Nao robot with the integrated Kinect successfully identifying landmarks of people positioned in front of it.

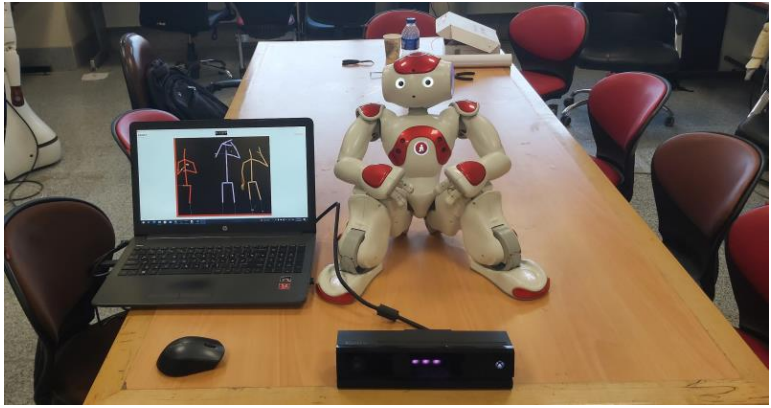

**Figure 8.** Nao Robot Utilizing Kinect for Precise Person Identification.

To assess the real-world performance of our model, we provided two 2-minute videos showcasing the robot's performances in different social situation scenarios. These videos were based on the best models (i.e., the highest accuracy) derived from either a 2D animation dataset or a virtual reality dataset. A study was conducted involving 36 new participants who were invited to view these videos via social media platforms. The participants were asked to rate their agreement with statements provided in **Table 1** on a 5-point scale, ranging from 'totally agree' to 'totally disagree', to assess the performance of each video. To eliminate potential biases, the order of the videos and questions was shuffled (counterbalance condition). Importantly, the participants were unaware of which video corresponded to which gaze patterns. To investigate whether the independent factors "Model (includes two levels: 2D or 3D animation)" and "Robotics Knowledge (includes two levels: Experts or Novices)" and their interaction have a statistically significant effect on the "Statements' Score" of the participants, two-way mixed ANOVA analysis were performed. The mentioned results are presented in **Table 2**. As can be seen in **Table 2**, no interactions were observed between the considered independent factors.

The results revealed that statements 1 and 2, relating to overall satisfaction and perception of the robot's interaction, received scores ranging from 3.5 to 4.2 for both models. These scores indicated that participants leaned towards agreement, rating them as 'agree'. Novices scored lower than experts. Furthermore, all groups rated the model trained with virtual reality higher. The results of the two-way mixed ANOVA show that there are no significant differences in the performance of the two models. However, for the first sentence, there is a significant difference in scores related to the knowledge of the participants about robots.

Statements 3, 4, and 5, which related to the robot's human-like qualities and realism, received scores around 3, indicating a neutral stance among participants, with experts scoring slightly higher. These scores suggest that participants found it challenging to fully accept the robot as a social companion or view it as human. The two-way mixed ANOVA results revealed significant differences in the performance of the two models for Statement 4. Additionally, there were significant differences in scores based on participants' knowledge about robots across nearly all statements.

Statements 6 and 7, regarding the attention and intelligence of the robot, received scores more than 3.5, indicating that participants mostly agreed with these statements. Additionally, participants believed that both models demonstrated equal intelligence. The two-way mixed ANOVA results indicate that there are no significant differences in the performance of the two models, as well as no significant differences in scores based on the participants' knowledge about robots.

Finally, statements 8, 9, and 10, which related to the responsiveness and understanding of human actions, received scores between 3 and 4. These scores indicated that participants generally agreed with these statements. Experts scored higher than the other two groups in this aspect. The two-way mixed ANOVA results show no significant differences in the performance of the two models. However, there are significant differences in scores for two statements for participants' knowledge about robots.

**Table 2.** The evaluation results obtained from the questionnaire and the analysis of variance (The p-values that are less than 0.05 are shown in bold).

| Statement number | 2D Animation | | Virtual Reality | | P-Value | | |
|---|---|---|---|---|---|---|---|
| | Novices | Experts | Novices | Experts | Factor 1: Model | Factor2: Knowledge | Interactions of factors |
| Statement 1 | 3.54 (0.87) | 4.08 (0.86) | 3.67 (0.62) | 4.17 (0.8) | 0.501 | **0.036** | 0.893 |
| Statement 2 | 3.46 (0.91) | 3.83 (0.99) | 3.67 (0.69) | 4.08 (0.76) | 0.192 | 0.121 | 0.904 |
| Statement 3 | 2.79 (0.91) | 3.5 (1.26) | 2.96 (0.98) | 3.58 (1.11) | 0.486 | 0.052 | 0.816 |
| Statement 4 | 2.46 (0.76) | 3.33 (1.43) | 2.88 (0.78) | 3.83 (1.21) | **0.006** | **0.008** | 0.792 |
| Statement 5 | 2.75 (0.92) | 3.58 (1.26) | 3.08 (0.91) | 3.67 (1.18) | 0.172 | **0.046** | 0.408 |
| Statement 6 | 3.88 (0.73) | 4.08 (0.86) | 3.83 (0.62) | 4.17 (0.8) | 0.897 | 0.211 | 0.699 |
| Statement 7 | 3.46 (0.91) | 3.92 (0.86) | 3.5 (0.71) | 3.92 (1.04) | 0.887 | 0.126 | 0.887 |
| Statement 8 | 3.33 (0.85) | 3.83 (1.07) | 3.46 (0.91) | 4.17 (0.99) | 0.212 | **0.043** | 0.567 |
| Statement 9 | 3.67 (0.69) | 3.92 (0.95) | 3.67 (0.62) | 4.08 (0.76) | 0.605 | 0.127 | 0.605 |
| Statement 10 | 3.25 (0.88) | 4.0 (1.0) | 3.42 (0.7) | 4.0 (0.82) | 0.672 | **0.007** | 0.672 |

**Figure 9-a** depicts a frame from the recorded video showcasing the implementation of the model trained with the 2D animation dataset on the Nao robot. The frame captures three individuals, one in close proximity and two at a distance. Person 1 and Person 2 can be seen standing, while Person 3 is nearer and waving. The Nao robot directs its head towards Person 3. In **Figure 9-b**, another frame from the recorded video illustrates the implementation of the model trained with the virtual reality dataset. In this frame, Person 1 is closer to the Nao robot and has crossed arms on the chest. Person 2 and Person 3 are both at a distance, but Person 2 also has crossed arms on the chest while Person 3 is pointing at Person 1. Nao's gaze in this frame is focused on Person 1.

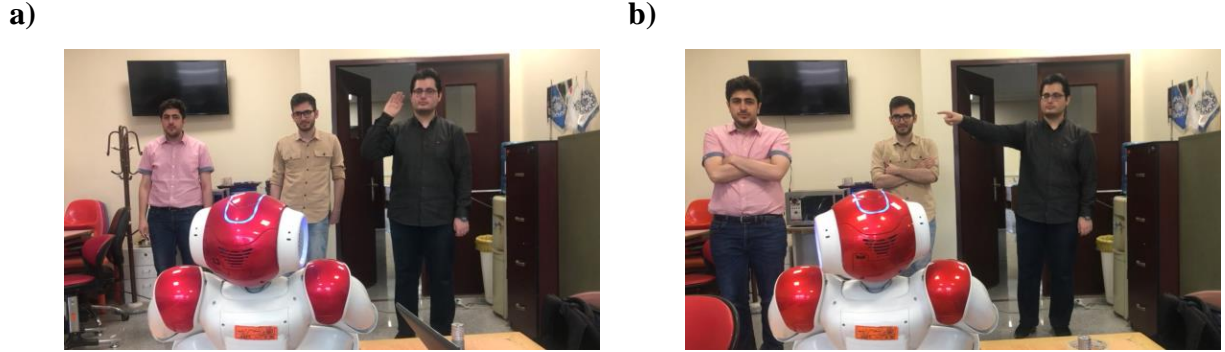

**Figure 9.** A frame of the recored video of the performance of the model on the Nao robot trained with **a)** 2D animation and **b)** 3D animation.

## 5. DISCUSSION

The findings of this study contribute to the growing body of literature on human-robot interaction, particularly in the area of gaze control systems for social robots. Our results demonstrate that the use of neural network architectures, such as LSTM and Transformers, enhances the accuracy of gaze prediction in social scenarios. This is consistent with previous research that highlights the importance of leveraging deep learning techniques to model complex human behaviors in robotic systems.

To contextualize and evaluate our work in relation to other related research, we sought to compare our results with those obtained by Mashaghi et al. which has been conducted on the same 2D animation dataset (however with other classical machine learning algorithms) [33]. They proposed that the gaze behavior can be modeled with equation (1), where $\prod w_i$ represents social stimuli; That is, each social stimulus that a person is doing has a weight, which is denoted by $w_i$. The multiplication of these weights has a direct effect on the level of attracting a person's attention. The expressions $P(r)$ and $\Theta(\theta)$ also show the effect of distance and angle in the amount of attracting attention, respectively. They trained their model by optimizing parameters using genetic algorithm. Their model has an accuracy of 51.3%.

$$\text{EA} = \left(\prod w_i\right) P(r) \Theta(\theta) \tag{1}$$

Another work that we compared our results was the work which was done with Aliasghari et al. [32]. They found that the gaze behavior can be modeled with equation (2), where $w_k$ is the weight factor corresponding to each social cue and n is the total number of social cues of the *i* th person. Proxemics (P) and Orientation (O) coefficients adjust the importance of all social cues existing in each individual based on his/her distance and angle relative to the robot. Their model considers the cue of talking, waving, entering, and leaving and also coefficients of Proxemics and Orientation. Their model reached an accuracy of 47.2% (on our 2D animation dataset considering their coefficients).

$$\text{EA} = \left(\sum_{k=1}^{n} w_k\right) P(r) O(\theta) \qquad (2)$$

Duque-Domingo et al. [18] introduced a competitive neural network approach that processes multiple stimuli, such as gaze, speech, and movement, to dynamically determine the robot's focus in multi-person interactions, our research extends these concepts by employing advanced neural network architectures, such as LSTM and Transformers, specifically tailored to adapt to complex social cues in real-time. Additionally, while Duque-Domingo's experiments were centered around static multi-person scenarios, our work further delves into dynamic environments, offering broader applicability in diverse social settings.

Interestingly, our results also reveal that participants with prior robotics experience rated the robot's performance more favorably, particularly in terms of its intelligence and responsiveness. This aligns with the notion that individuals familiar with robotic systems have more realistic expectations and are better able to appreciate the capabilities of advanced gaze control algorithms. However, novices, influenced by portrayals of robots in media, expected more human-like qualities from the robots, indicating a gap between public perception and current technological capabilities [34], [35].

Overall, this study provides a valuable proof of concept for the use of deep learning in enhancing the social interactions of robots through improved gaze control. By bridging the gap between human expectations and robotic capabilities, these advancements could pave the way for more intuitive and effective human-robot collaborations in various settings.

## 6. LIMITATIONS AND FUTURE WORK

One of the main limitations of this study was the small number of participants from whom we extracted the gaze models. While increasing the number of participants would undoubtedly result in a more generalizable model, the observed results based on the robot's performance were fairly promising. This indicates that our extracted models can be considered as a proof of concept for providing gaze control systems for social robots. The results of the two studies revealed that one of the main challenges lies in the fact that determining the specific location a person should look at is not always straightforward. It can vary depending on individual preferences and cultural differences, thereby introducing additional complexity to the gaze control system. The number of social states is countless, and only a small number of them were considered in this experiment. Social situations such as the gender of people, the color of clothes, and the volume of the voice can all affect the type of look. The conducted research encountered challenges that affected the accuracy of the designed models, and some of them are mentioned here. In the first experiment, the duration of the animation was 11 minutes, and data collection for this period of time was done by the tracker. Fixed eye position was maintained for a long time, causing participants to become tired after a few minutes. As a result, the accuracy of the data collected in the last minute could be compromised. Additionally, the designed animation did not contain many social situations and only had animated stimuli. In the second experiment, an animation with more social situations and activities was used to address the challenges faced in previous studies. However, the main challenge encountered in this research was that the virtual reality glasses were not equipped with an eye tracker. Consequently, during the data collection, participants had to turn their heads and look only straight ahead with their eyes. Future work can address these challenges by equipping virtual reality glasses with eye trackers and also face the challenge of gathering data for model training with new deep machine learning algorithms.

## 7. CONCLUSIONS

This study explored the effectiveness of LSTM and Transformer neural networks in predicting gaze behavior for social robots during a preliminary exploratory study. Across two preliminary basic experiments, our models demonstrated consistent accuracy, with improvements noted when the models were allowed more flexibility in selecting the correct gaze target. Implementing these models on a NAO robot showed that participants appreciated the robot's responsiveness and perceived intelligence, though they did not view the robot as a social companion. Notably, participants with prior robotics experience rated the robot's performance more favorably, while novices had higher expectations regarding the robot's human-like qualities. Despite some challenges in achieving high accuracy, this research provides valuable insights into the development of gaze control systems for social robots, laying the groundwork for future enhancements in human-robot interaction.

**Acknowledgment**

This study was funded by the "Dr. Ali Akbar Siassi Memorial Research Grant Award" and Sharif University of Technology.

**Conflict of interest**

Author Alireza Taheri has received a research grant from the Sharif University of Technology. The author Ramtin Tabatabei asserts that he has no conflict of interest.

**Availability of data and material (data transparency)**

All data from this study is available in the archive of the Social & Cognitive Robotics Laboratory.

**Code availability:**

All of the codes are available in the archive of the Social & Cognitive Robotics Laboratory. In case the readers need the codes, they may contact the corresponding author.

**Authors' contributions:**

Both of the authors contributed to the study conception and design. The idea of the study was presented by Alireza Taheri. Material preparation, data collection, and analysis were performed by Ramtin Tabatabaei. Alireza Taheri supervised the study. The first draft of the manuscript was written by Ramtin Tabatabei; and both authors commented on previous versions of the manuscript. Both authors read and approved the final manuscript.

**Consent to participate:**

Informed consent was obtained from all individual participants included in the study.

**Consent for publication:**

All of the participants have consented to the submission of the results of this study to the journal.